\documentclass[manuscript,acmsmall,nonacm,screen]{acmart}

\usepackage[T1]{fontenc}
\usepackage[utf8]{inputenc}
\usepackage{amsmath,amsfonts}
\usepackage{graphicx}
\usepackage{array}
\usepackage{academicons}
\usepackage{xcolor}
\usepackage{stfloats}
\usepackage{url}
\usepackage{verbatim}
\usepackage{xspace}
\usepackage{comment}
\usepackage{csquotes}
\usepackage{multirow}

\usepackage{amsmath,amsfonts}
\usepackage{amsthm}
\usepackage{bm}

\usepackage{fancyhdr}

\usepackage{amsthm} % <-- Add this line to use theorem environments
\usepackage{algorithm}

\usepackage{algpseudocode}
\usepackage{textcomp}
\usepackage{listings}
\usepackage{color}
\usepackage{wrapfig}

\usepackage{soul}
\usepackage{xcolor}

\usepackage[most]{tcolorbox} % color box

\usepackage{adjustbox}
\usepackage{balance}

\usepackage{array}
\usepackage{ragged2e}
\newcolumntype{P}[1]{>{\RaggedRight\hspace{0pt}}p{#1}}
\newcolumntype{X}[1]{>{\RaggedRight\hspace*{0pt}}p{#1}}
\newcolumntype{C}[1]{>{\centering\arraybackslash\hspace{0pt}}p{#1}}

\usepackage{orcidlink}

\usepackage{xspace}

\usepackage{amsmath,amsfonts}

\usepackage{soul}
\usepackage{balance}
\usepackage{hyperref}
\usepackage{textcomp}
\usepackage{listings}

\definecolor{codegreen}{rgb}{0,0.6,0}
\definecolor{codegray}{rgb}{0.5,0.5,0.5}
\definecolor{codepurple}{rgb}{0.58,0,0.82}
\definecolor{backcolour}{rgb}{0.95,0.95,0.92}

\usepackage{graphicx} 
\usepackage{caption,subcaption}
\usepackage{csquotes}

\usepackage{pdfcolfoot}

\usepackage{enumitem} % better itemize

\newcommand{\eg}{{e.g.,}\xspace}
\newcommand{\viz}{{viz.,}\xspace}

\newcommand{\etal}{{et~al.}\xspace}
\newcommand{\ie}{{i.e.,}\xspace}

\newcommand{\ci}{{\it (i) }}
\newcommand{\cii}{{\it (ii) }}
\newcommand{\ciii}{{\it (iii) }}
\newcommand{\civ}{{\it (iv) }}

\newcommand{\ca}{{\it (a) }}
\newcommand{\cb}{{\it (b) }}
\newcommand{\cc}{{\it (c) }}
\newcommand{\cd}{{\it (d) }}

\newcommand{\cpar}[1]{{\vspace*{0.5\baselineskip}\noindent\bfseries #1\quad}}

\definecolor{notecolor}{rgb}{0.8,0,0} % A darker red

\newcommand{\mhtodo}[1]{{\hl{\textbf{TODO}}~\color{blue}\textbf{\textit{#1}} \textbf{--- MH}}}

\theoremstyle{definition}
\definecolor{scol}{rgb}{0, 0, 1}

\newcommand*\circled[1]{{\small\tikz[baseline=(char.base)]{
            \node[shape=circle,draw,inner sep=1pt] (char) {#1};}}}

\newcommand{\ttt}[1]{\texttt{#1}}

\AtBeginDocument{
	
	\captionsetup[figure]{belowskip=0pt, aboveskip=5pt}
	\captionsetup[table]{belowskip=0pt, aboveskip=5pt} 
	
	\setlength{\abovecaptionskip}{0pt}    %default 10pt
	\setlength{\belowcaptionskip}{0pt}    %default 0pt

    \setlength\textfloatsep{0.75\baselineskip}

}

\setcopyright{none}
\renewcommand\footnotetextcopyrightpermission[1]{}
\begin{document}

% {\pagestyle{empty}
% \setcounter{page}{1}
% \input{coverletter}
% \newpage
% \setcounter{page}{1}
% }

% \title{Detecting Data Inconsistencies in Agricultural Weather Networks with Digital Twins}

\title{Weather-Driven Agricultural Decision-Making Using Digital Twins Under Imperfect Conditions}

\author{Tamim Ahmed}
\email{tamim.ahmed@wsu.edu}
\orcid{1234-5678-9012}
\affiliation{%
  \institution{Washington State University}
  % \city{Pullman}
  % \state{Washington}
  \country{USA}
  % \country{} 
}

\author{Monowar Hasan}
\email{monowar.hasan@wsu.edu}
\orcid{1234-5678-9012}
\affiliation{%
  \institution{Washington State University}
  % \city{Pullman}
  % \state{Washington}
  \country{USA}
  %\country{} 
}

% paper name
\newcommand{\fancyname}{\textsc{Cerealia}\xspace}

\begin{abstract}
By offering a dynamic, real-time virtual representation of physical systems, digital twin technology can enhance data-driven decision-making in digital agriculture. Our research shows how digital twins are useful for detecting inconsistencies in agricultural weather data measurements, which are key attributes for various agricultural decision-making and automation tasks. We develop a modular framework named \fancyname that allows end-users to check for data inconsistencies when perfect weather feeds are unavailable. \fancyname uses neural network models to check anomalies and aids end-users in informed decision-making. We develop a prototype of \fancyname using the NVIDIA Jetson Orin platform and test it with an operational weather network established in a commercial orchard as well as publicly available weather datasets.
\end{abstract}




% \keywords{Digital Twin, Machine Learning, Weather Networks}

\maketitle

% \begin{center}
% \noindent\fbox{%
%     \parbox{0.98\columnwidth}{%
%     	\footnotesize
%         \textbf{Disclaimer.}\quad This research uses weather stations installed in a commercial orchard maintained by Washington State University, United States. While this is a closed system, access can be made available for research purposes by request, as we did in this work. We use this platform to demonstrate the efficacy of our approach in a real, live setup.  Our use of this platform, however, by no means implies our affiliation with this infrastructure/institution.
%     }%
% }
% \end{center}

\section{Introduction}

Modern agriculture, especially high-value specialty crop production, relies heavily on data-driven automation technologies for production management. Agricultural decision-making increasingly relies on precise and consistent environmental data to optimize crop health, resource utilization, and yield predictions. Regional as well as localized weather critically drives several on-farm decision-support models and intelligent automated systems, including irrigation, abiotic stress (e.g., cold, heat) mitigation, and pest management, among others~\cite{boomgard2022machine}.  Weather data measurements, which include essential metrics such as temperature, humidity, and precipitation, form the foundation of such decisions. However, weather networks, like any cyber-physical and Internet-of-Things (IoT) systems, are often plagued by \textit{inconsistencies}\footnote{In this work, we focus on \textit{effects} on inconsistent/irregular weather streams rather than the \textit{causes} that trigger the anomalies.} due to sensor faults, calibration drift, communication errors, environmental interference, or cyber breaches. Imperfect measurements can significantly undermine the reliability of downstream applications in agriculture, ranging from predictive modeling to real-time interventions. Such discrepancies can lead to inaccurate predictions, suboptimal resource allocation, and reduced agricultural productivity.

 Any inconsistent weather data input could lead to inadequate decision-making and wrongful actuation of precision management scenarios, resulting in considerable crop yield and quality losses. To address this problem, we adopt the \textit{digital twin} technology~\cite{barricelli2019survey}
 %we build a digital twin platform 
 for analyzing and understanding agricultural weather networks that are critical for grower decision support. We build dynamic virtual data streams that mirror physical stations---to monitor, validate, and enhance the reliability of weather data in the face of anomalies. While digital twin technology is used in other cyber-physical domains (\eg control systems, manufacturing, robotics), its benefits are not fully explored for digital agriculture.  By leveraging the concept of digital twins, we show that it is feasible to create a virtual replica of weather networks for real-time inconsistency detection, data imputation, and predictive analytics. We name our framework \fancyname.\footnote{In ancient Roman religion, the Cerealia was a major festival dedicated to the sowing, growing, and harvesting of crops. As our work aims to assist in better decision-making for agricultural stakeholders, we name our framework \fancyname.}

\begin{figure}[!t]
\centering
\includegraphics[scale=0.4]{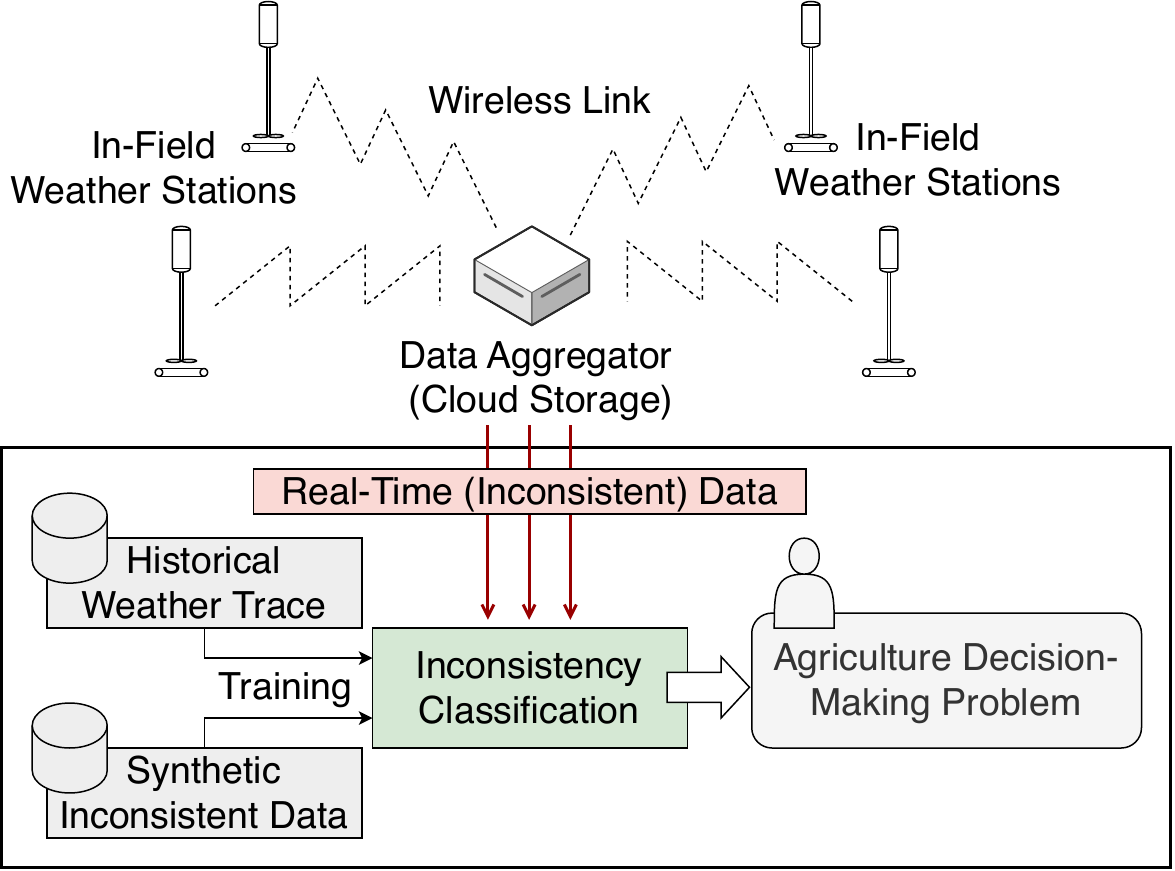}
\caption{High-level schematic of this work. \fancyname stores historical weather traces and simulated anomalous data that is used to detect runtime inconsistent data generated by the remote weather stations deployed in the field. The end users can use \fancyname to analyze how inconsistent data impacts the target agriculture decision-making process.}
\label{fig:overview}
\end{figure}

 Figure~\ref{fig:overview} presents a high-level overview of \fancyname. The ``twin'' of weather stations stores a historical database of past weather measurements and uses a synthetic data generator engine that is capable of emulating ``noisy'' (\viz inconsistent) data. Based on these observations, \fancyname uses machine learning models to check for inconsistencies in the real-time weather feed coming from the in-field weather stations. Our current integration of \fancyname integrates nine neural network-based machine learning models (see \S\ref{sec:ml}). These models are optimized for detecting and addressing inconsistencies in weather data.
 
 A key feature of \fancyname is to design it as a \textit{modular tool} that end-users can utilize for several in-farm decision-making tasks when perfect physical measurements are not available.  As a user-inspired scenario, we study inconsistency implications, \ie how erroneous weather data impacts in-farm decision-making. Specifically, we conduct \textit{two comprehensive case studies} that demonstrate its applicability and impact.  Our first case study \textit{predicts weather data}. This is useful for weather data imputation applications, where growers are required to predict missing or inconsistent weather data, perhaps due to system malfunction or lack of stations in the target regions. The second case study involves \textit{predicting fruit surface temperature}, which is a crucial indicator of fruit stress. Growers use temperature prediction tools to initiate automated cooling precautions to reduce crop losses. We show that \fancyname is useful to observe whether inconsistent data affects the surface temperature prediction.

 %released by the Max Planck Institute~\cite{jena_weather_dataset}.

\cpar{Our Contributions.}
By bridging computing and agriculture, \fancyname represents a significant step toward building resilient, data-driven systems capable of withstanding the uncertainties in weather measurements. The contributions of this paper are as follows:

\begin{itemize}
	\item We introduce \fancyname, a digital twin-based framework that addresses data inconsistencies using neural network models in agricultural weather networks [\S\ref{sec:design-workflow}].
	\item By leveraging \fancyname, we demonstrate the usefulness of digital twins for agricultural decision-making using two realistic case studies (predicting missing weather data and fruit surface temperatures) [\S\ref{sec:casestudy}].
\end{itemize}

\fancyname is implemented and tested by using a commercial orchard located in Central Washington, United States (\S\ref{sec:impl}--\S\ref{sec:eval}).\footnote{This research uses weather stations installed in a commercial orchard maintained by Washington State University AgWeatherNet team~\cite{awn}. While this is a closed system, access can be made available for research purposes by request, as we did in this work. We use this platform to demonstrate the efficacy of our approach in a real, live setup.  Our use of this platform, however, by no means implies our affiliation with this infrastructure/institution.} The digital twin modules are deployed on an NVIDIA Jetson Orin platform~\cite{karumbunathan2022nvidia}. In addition to live weather feed from the orchard, \fancyname is also tested using a publicly available European weather dataset~\cite{jena_weather_dataset}. We conduct a comprehensive evaluation of nine state-of-the-art neural network models to analyze their effectiveness in detecting weather data inconsistencies. We measure the overheads of running learning models on our evaluation platform and test the scalability of \fancyname. We find that the overheads of checking (in terms of timing and memory usage) are minimal (less than 1~s and 0.5~MB of memory use).  Our research highlights the potential of digital twin tools like \fancyname to improve agricultural decision-making under imperfect conditions. % contributing to more sustainable and efficient farming practices.

\begin{center}
\noindent\fbox{%
    \parbox{0.98\columnwidth}{%
    \fancyname implementation is publicly available: {\tt\url{https://github.com/CPS2RL/ag-dt}}.
    }%
}
\end{center}

\section{Background and Context}

We now start with a background on weather implications for agriculture and the rationale of this work (\S\ref{sec:bg_weathernet}). We then provide a brief background on how a digital twin architecture like \fancyname can aid the weather-driven agricultural decision-making process (\S\ref{sec:bg_dt}).

\subsection{Weather Data Impacts on Agriculture} \label{sec:bg_weathernet}
 
Weather data has a significant impact on agriculture. For example, in early spring and late winter, frost damage in fruit crops can hurt fruit production. Fruit growers attempt to mitigate frost using wind machines, propane heaters, and under-tree sprinklers~\cite{evans1999frost}. To activate those mitigation tools, growers use weather data-driven models and sensing systems that automate and improve frost management with resource conservation (natural gas/energy)~\cite{talsma2023frost,cann2022observing}. Similar to frost,  extreme and more frequent heat stress events in summer can cause fruit cullage due to heat stress instigated by sunburn and related damages. To address this problem, researchers have explored weather-driven models to increase the effectiveness of heat stress mitigation techniques~\cite{goosman2023apple,ranjan2020field,amogi2023mask}. In addition to frost and heat management, uninterrupted and reliable weather data obtained at the farm scale is necessary for several real-time tasks, including irrigation, disease and pest management, fruit growth prediction, among others~\cite{boomgard2022machine}.

Existing weather-driven agriculture decision-making models~\cite{talsma2023frost,cann2022observing, goosman2023apple,ranjan2020field,saxena2023grape,amogi2023mask,amogi2024edge} rely on the availability of ``perfect'' weather information. Like any networked cyber-physical system, weather stations are also susceptible to faults and cyberattacks~\cite{boomgard2022machine, weiwei2018fault, botschner2024cybersecurity, kulkarni2024review}. Inadequate and inconsistent environmental data, for instance, due to faults in weather station sensors or adversarial breaches, often fail to protect the crop during extreme heat or cold events. Further, any unreliable decision-making model or lack of automation requires the growers to work on the farm in extreme weather conditions to manually activate protection measures (e.g., turning on heathers, putting protective netting, or tuning on the cooling system to prevent heat stress). We envision that a digital twin platform incorporating measurement inconsistencies provides a virtual view of the weather network and can help growers make informed decisions in their target agriculture applications. 

\subsection{Digital Twin and its Usefulness in Agriculture} \label{sec:bg_dt}

In literature, a digital twin is defined as ``a digital replica of a living or non-living physical entity''~\cite{el2018digital}. Digital twin technology is used to describe and model the corresponding physical counterpart digitally by combining modeling, simulation, and emulation technologies with other analytics to better understand aspects of the target application domain. Digital twins are commonly used in many cyber-physical domains such as manufacturing~\cite{kritzinger2018digital}, IoT~\cite{minerva2020digital}, healthcare~\cite{sun2023digital}, and recently, digital agriculture~\cite{purcell2023digital}. However, its efficacy in weather-driven agricultural decision-making still has not been explored. 

To realize the benefits and limitations of existing weather-based agriculture models in the presence of inconsistent data, we need a realistic ``virtual'' platform to mimic data impurities and study how it impacts agriculture decision-making (for instance, predicting the surface temperature of fruits). This research is the first step to building a digital twin of an agricultural weather network that models data inconsistency. Our proposed framework, \fancyname, allows us to detect imperfect weather measurements using both open-field mesoscale and synthetic data. Besides, \fancyname helps end-users analyze the impacts of imperfect data on agricultural decision-making. Another benefit of \fancyname is that it allows us to create a virtual weather ecosystem that connects with physical stations and can synthetically produce noisy data without tampering with in-field stations. As a result, we can artificially recreate inconsistent behaviors and analyze their impact without causing any service disruptions in weather network operations. 

As we shall see in this paper, \fancyname combines historical data and synthetically generated anomalies to detect inconsistencies in real-time weather feeds. \fancyname does so by incorporating several machine learning models. We perform two concrete case studies to realize further how \fancyname can be useful for agriculture stakeholders. Our first case study (see \S\ref{sec:cs_impute}), \textit{weather data imputation problem}, predicts various weather parameters (\eg temperature, pressure, radiation, wind). These imputed values could be useful when measurements from all stations/sensors are unavailable or as a ground truth of a mesonet dataset for data-driven analytics.  The second case study (see \S\ref{sec:cs_fst}) considers the \textit{fruit surface temperature prediction} problem and quantitatively analyzes whether inconsistent data results in accuracy loss in predicting apple surface temperature.

 \subsection{The Novelty of this Research} 
 
 This is an interdisciplinary research where the core is an IoT system (weather network) that is used for domain-specific (\viz agriculture) data analytics facilitated by digital twin technology. 
 %As mentioned in \S\ref{sec:bg_weathernet}, although biosystems researchers develop some models for agricultural decision-making tasks, they assume a perfect setup in which sensor readings are fault/attack-free, which is not the case in practice. For realistic evaluation and correct decision-making, we need to filter out anomalies. 
 As mentioned in \S\ref{sec:bg_weathernet}, although biosystems researchers develop models for agricultural decision-making tasks, they often assume a perfect setup in which sensor readings are free from faults or attacks---an assumption that does not hold in practice. Our work leverages the power of digital twins to bridge the gap between the physical weather network and data-driven agricultural decision-making in the presence of imperfect measurements. We note that time series prediction and anomaly detection are not new research areas. However, our contribution is \textit{not} on proposing a new detection or prediction algorithm. Instead, we introduce a \textit{modular} design that enables ag-tech stakeholders to: \ca use any learning tools to analyze inconsistencies in measurements and \cb evaluate how these inconsistencies impact decision-making activities (e.g., predicting the temperature of fruits).  With a tool like \fancyname, end-users can analyze inconsistencies in weather data and their impact on agricultural decision-making in real time. Our modular design allows seamless integration of both new and existing machine learning and decision-making models.

\begin{figure}[!t]
\centering
\includegraphics[scale=0.45]{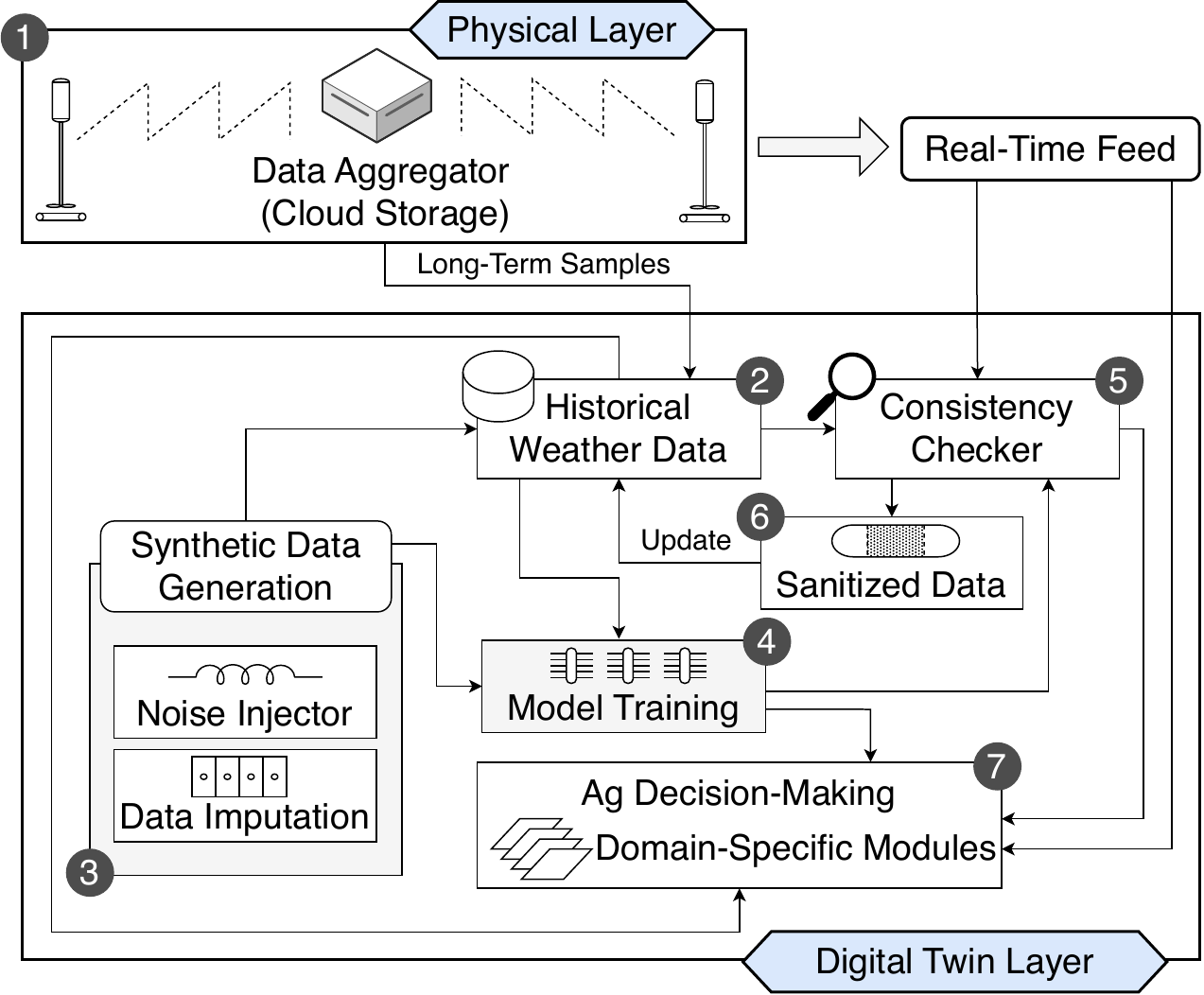}
\caption{Workflow of \fancyname. We make \fancyname modular that allows designers to integrate various noisy data that can be used with historical weather traces to train machine learning model(s). A runtime consistency checker module uses a trained model to check for imperfect measurements and their impact on targeted decision-making applications (for instance, fruit heat/frost prediction).}
\label{fig:pipeline}
\end{figure}

\section{Design and Workflow} \label{sec:design-workflow}

The high-level architecture of \fancyname is illustrated in Fig.~\ref{fig:pipeline}. We design \fancyname using a two-layer architecture: \ca \textit{physical} layer and \cb \textit{digital twin} layer, as we present below.

\subsection{Components of \fancyname} \label{sec:components}

\subsubsection{Physical Layer} 

As shown in Block \circled{1} of Fig.~\ref{fig:pipeline}, the physical layer consists of weather stations deployed in the field. Many off-the-shelf weather stations (\eg Cabled Vantage Pro2~\cite{davisinstrumentsCabledVantage}, ATMOS 41W~\cite{metergroupATMOSMETER}) and existing weather networks (\eg AgWeatherNet~\cite{awn}) generally collect field measurements from stations in a periodic interval and store them in a cloud gateway for further processing. We follow a similar approach in \fancyname, where a cloud storage service collects data from distributed stations and feeds this to our analysis engines in the digital twin layer (see next).

\subsubsection{Digital Twin Layer} 

The digital twin layer is the crux of the \fancyname and contains several modules (Blocks~\circled{2}-\circled{7} in Fig.~\ref{fig:pipeline}). %This layer operates in two phases. 
In the \textit{offline phase} (gray-shaded blocks in the Digital Twin layer), \fancyname collects historical field measurements accumulated over the years from stations of interest or existing region-specific weather datasets and stores them in its internal database (Block~\circled{2} in the figure). These historical traces could be sanitized   (\ie free from anomalous data) or may have inconsistent measurements. As the focus of \fancyname is to emulate various inconsistent scenarios and see how they affect application-specific decision-making, we incorporate a noise generator module (Block~\circled{3}) that allows us to inject various types of anomalous data. Our current setup considers four types of inconsistencies. They are elaborated in \S\ref{sec:noise}. We train machine learning models based on historical observations and noise traces to check for inconsistent data at runtime (Block~\circled{4}). \fancyname currently tested with nine state-of-the-art machine learning models, which we discuss further in \S\ref{sec:ml}. 
%These offline models are used in the runtime phase of \fancyname to aid agriculture decision-making.

The \textit{runtime} components of \fancyname (Blocks~\circled{5}-\circled{7}) use trained models and real-time measurements obtained from the physical layer to check for data inconsistencies. If an anomalous data is detected, \fancyname notifies the user. Further, the historical database can be further updated by using sanitized or imputed data (Block~\circled{6}) from the consistency checker (Block~\circled{5}). Besides, the users can use the trained models and real-time measurements to see how noisy data affects the prediction of a given agricultural task (Block~\circled{7}). We discuss this further in \S\ref{sec:desmak} and perform two agriculture case studies (see \S\ref{sec:casestudy}) to demonstrate how \fancyname can be useful in agricultural prediction tasks. 

\fancyname supports \textit{incremental} learning and updates. This allows \fancyname to overfit and adapt for the targeted weather network, which is often the key requirement for regional agricultural decision-making. For instance, as new samples are generated, \fancyname updates the historical database and retrains the model with new observations (the circular loops in Blocks~\circled{5}, \circled{6}, \circled{2}, \circled{4}). As model training typically takes time and weather data is generated at a higher volume/frequency, designers can opt for model retraining at a coarse granularity. The granularity of updating the historical database and the retraining models is left as a designer-chosen parameter. Our evaluation considers an on-shot scenario (\ie without any retraining) and demonstrates the performance of inference and effectiveness of runtime decision-making, as model training is typically conducted offline (\ie when the system is not operational).

The key components of the digital twin layer (\ie Blocks~\circled{3}-\circled{7}) are modular by design. As mentioned above, our current setup incorporates four kinds of noisy data and uses nine machine-learning models to see the models' behavior under these inconsistencies to find the best possible solution for \fancyname. However,  other faulty/anomalous data and other statistical models or anomaly detection tools can be incorporated with \fancyname to check for inconsistencies. Likewise, as discussed in \S\ref{sec:casestudy}, for demonstration purposes, we tested \fancyname for two agricultural use cases (\eg weather data imputation and fruit surface temperature prediction problems). However, this component of \fancyname (\ie Block~\circled{7}) can be adapted for any other agriculture decision-making tasks (\eg fruit bloom phenology~\cite{chaves2015modeling}, soil water content prediction~\cite{nath2024soil}) without loss of generality.

% \hl{Here, we tried to address a key requirement of digital twin system adaptability. We have introduced an incremental training loop which will allow models} (Blocks~\circled{4}) \hl{to continuously update using the real-time measurement from the physical layer. It can ensure that the system is not only analyzing data but also adapts to it.} 

\begin{figure}[!t]
\centering
\includegraphics[scale=0.12]{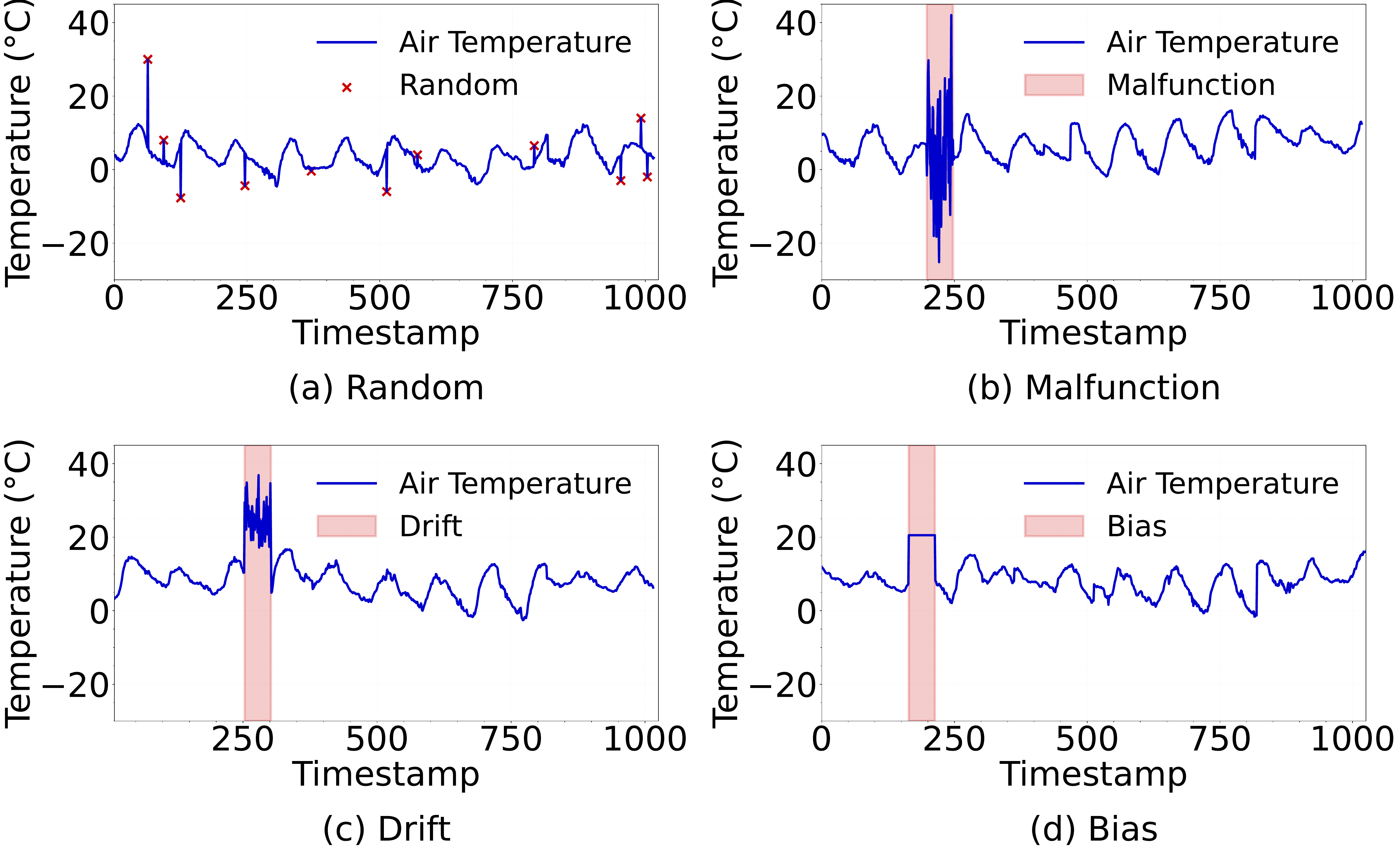}
\caption{Visualization of the data inconsistencies incorporated with \fancyname. The plot shows temperature readings over time in the presence of the following inconsistencies: \ca random (top-left), \cb malfunction (top-right), \cc drift (bottom-left), and \cd bias (bottom-right).}
\label{fig:faults}
\end{figure}

\subsection{Generating Noisy Measurements} \label{sec:noise}

One of the benefits of having a digital twin architecture like \fancyname is that we can emulate the behaviors of actual weather stations with customized parameters. 
% At each time instance \( t \), each station captures a set of readings from $n$ sensors. Let us represent these readings as follows:
%     \begin{equation}
%         \mathbf{X}(t) = \{ x_1(t), x_2(t), \dots, x_n(t) \},
%     \end{equation}
%     where \( x_i(t) \) represents the recorded value for weather attribute \( i \) (\eg air temperature, atmospheric pressure, solar radiation, wind speed). As mentioned in \S\ref{sec:components}, \fancyname maintains a database of historical weather traces. Let us denote the historical dataset as follows:
%     \begin{equation}
%         H = \{ \mathbf{X}(t_1), \mathbf{X}(t_2), \dots, \mathbf{X}(t_m) \},
%     \end{equation}
%     where \( t_1, t_2, \dots, t_m \) are the timestamps of the recorded entries. 
%     \hl{TODO: need to check where to place -- MH.}
Although physical station measurements could be anomalous in practice, \ca they are unpredictable, and  \cb we need a reference model (\ie ground truth) to check for abnormalities. Hence, we use a noise generator~\cite{de2016benchmark} to mimic inconsistent behaviors.  We consider four kinds of inconsistent data: \ca random (random spikes in the measurements), \cb malfunction (abrupt oscillations in the measurements), \cc drift (similar to malfunction, but the results are significantly shifted from nominal values), and \cd bias (results are scaled up or down than normal values). Figure~\ref{fig:faults} illustrates these four kinds of faults. Although we choose these four kinds of inconsistencies for our current demonstration---as they often represent typical sensor faults~\cite{ni2009sensor}---any other fault model can be integrated with \fancyname's synthetic data generator module (\ie Block \circled{3} in Fig.~\ref{fig:pipeline}), which then will be fed into the inconsistency checker (\ie Block \circled{5} in Fig.~\ref{fig:pipeline}, discussed next).

% \fancyname maintains \textit{tags} to label clean and anomalous data as follows:  $ \mathcal{C} = \{\mathtt{clean}, \mathtt{random}, \mathtt{malfunction}, \mathtt{drift}, \mathtt{bias}\}$, where $\mathtt{clean}$ tag denote no noise is injected in the data. The other tags, \ie $\mathtt{random}$,  $\mathtt{malfunction}$, $\mathtt{drift}$, and $\mathtt{bias}$ represent corresponding noise class, as presented below.

\subsection{Leveraging Machine Learning for Inconsistency Checking} \label{sec:ml}

Based on historical traces and a given noise model, we train machine learning models to characterize off-nominal data from real-time weather measurements generated from the physical layer. 
Some machine learning models, such as SVM~\cite{suthaharan2016support}, KNN~\cite{steinbach2009knn}, and K-Means~\cite{steinley2006k}, often assume static data relationships or require handcrafted features, which may not be efficient for capturing the complex sequential nature of weather measurements. Also, they often struggle with large datasets, particularly high-dimensional and noisy data (which is the case for weather networks).
Hence, we use deep learning models in \fancyname to check for inconsistencies.  
In our current implementation, we tested nine neural network models, which are discussed next. However, as mentioned earlier, other statistical or learning models can be integrated with \fancyname.

% \subsubsection{Ensemble Model} 

% Some machine learning models, such as SVM~\cite{suthaharan2016support}, KNN~\cite{steinbach2009knn}, K-Means~\cite{steinley2006k}, often assume static data relationships or require handcrafted features, which may not be efficient for capturing the complex sequential nature of weather measurements. Also, they often struggle with large datasets, particularly high dimensional and noisy data (which is the case for weather networks). Ensemble models such as Random Forest (RF)~\cite{breiman2001random}, CatBoost~\cite{prokhorenkova2018catboost} can capture complex features and nonlinear relationships for both regression and classifications~\cite{liaw2002classification}. However, ensemble methods cannot account for temporal dependencies, and thus, their effectiveness could be limited for sequential patterns. In our evaluation, we integrate one ensemble model (RF) into \fancyname to analyze whether the ensemble learning techniques are effective for weather data inconsistency checking.

\subsubsection{Convolutional Neural Network (CNN) Models} CNN models stand out in
retrieving hierarchical and local features from time-series data like weather measurements. We use two CNN models: \ca Temporal Convolution Network (TCN) \cite{he2019temporal} and \cb ResNet \cite{he2020resnet}. TCN uses causal convolution and dilation to capture short-term and long-term temporal dependencies. ResNet uses residual connections and expedites the training of deep networks with reduced training error, but often requires careful tuning due to its complexity.

\subsubsection{Recurrent Neural Network (RNN) Models} RNNs are highly effective for sequential data analysis due to their inherent memory capabilities. RNN models can capture the features and dependencies across long sequences. We incorporate the following RNN models for inconsistency checking: \ca Long Short-Term Memory (LSTM)~\cite{van2020review}, \cb Bi-LSTM~\cite{siami2019performance} and \cc Gated Recurrent Unit (GRU)~\cite{shewalkar2019performance}. LSTM is a well-known algorithm for retaining information over time. The latter two models are improved variants of LSTM that enhance sequence prediction.
%and \cc two Transformer-based models---Time-Series Transformer (TST)~\cite{zerveas2021transformer} and Informer~\cite{zhou2021informer} that are useful for capturing intricate patterns and long sequence in the weather station measurements.

\subsubsection{Transformer Models} Transformer architectures are useful in tracking relationships in sequential data. We use two transformer-based models: \ca Time-Series Transformer (TST)~\cite{zerveas2021transformer} and \cb Informer~\cite{zhou2021informer} that are useful for capturing intricate patterns and long sequences in the weather station measurements.

\subsubsection{Hybrid Models} We also use two hybrid models: \ca TST-AE~\cite{zhang2021unsupervised} and \cb LSTM-AE~\cite{provotar2019unsupervised}. TST-AE uses a transformer-based self-attention with an encoder-decoder structure for better feature extraction. LSTM-AE is based on an LSTM-based encoder-decoder and can provide efficient memory usage, but comes with the cost of reduced accuracy. 

%Appendix~\ref{?} provides additional details on how these models are configured in \fancyname.

% \subsection{Using \fancyname for Agricultural Decision-Making Tasks with Imperfect Measurements} \label{sec:desmak}

\subsection{Using \fancyname for Agricultural Decision-Making with Imperfect Measurements} \label{sec:desmak}

At each time instance \( t \), each station captures a set of readings from $n$ sensors. Let us represent these readings as follows:
    % \begin{equation}
        $\mathbf{X}(t) = \{ x_1(t), x_2(t), \dots, x_n(t) \}$,
    % \end{equation}
    where \( x_i(t) \) represents the recorded value for weather attribute \( i \) (\eg air temperature, atmospheric pressure, solar radiation, wind speed). As mentioned in \S\ref{sec:components}, \fancyname maintains a database of historical weather traces. Let us denote the historical dataset as follows:
    %\begin{equation}
        $H = \{ \mathbf{X}(t_1), \mathbf{X}(t_2), \dots, \mathbf{X}(t_m) \}$,
    %\end{equation}
    where \( t_1, t_2, \dots, t_m \) are the timestamps of the recorded entries. 
Note that this historical dataset $H$ does not necessarily contain all sanitized or perfect measurements. For a given time instance $t$, we mimic imperfect environments during model training by injecting noisy samples, $ \mathbf{X}_{noise}(t)$.  The formal description of $ \mathbf{X}_{noise}(t)$ for various noise classes is described in \S\ref{sec:inconsistency_generation}. \fancyname maintains \textit{tags} to label clean and anomalous data as follows:  $ \mathcal{C} = \{\mathtt{clean}, \mathtt{random}, \mathtt{malfunction}, \mathtt{drift}, \mathtt{bias}\}$, where $\mathtt{clean}$ tag denote no noise is injected in the data. The other tags, \ie $\mathtt{random}$,  $\mathtt{malfunction}$, $\mathtt{drift}$, and $\mathtt{bias}$ represent corresponding noise class.

% \fancyname uses the saved measurements $H$ to train a machine learning model, which is then used to check for inconsistencies at runtime. 
\fancyname allows the end-users to use the trained models (obtained from historical observations $H$, including inconsistent samples) and make informed decisions about their target applications. For instance, we can run an inference operation to check for inconsistent weather data at any time epoch $t$. 
% \fancyname uses trained models to detect inconsistencies in the weather station’s real-time feed. We 
\fancyname continuously fetches data from the physical layer and then classifies incoming sequences as clean or anomalous. Given a new real-time sequence of size $k$ \ie \( \mathbf{X} = \{ X(t - k), \dots, X(t) \} \), the model classifies it to detect any anomalies as follows:
    %\begin{equation}
    $y_{\text{real-time}} = f_{\text{pretrained\_model}}(\mathbf{X})$,
    %\end{equation}
    where $f_{\text{pretrained\_model}}(\mathbf{X})$ is the output (inference) from the machine learning model. If \( y_{\text{real-time}} \neq \mathtt{clean} \), the sequence is flagged as inconsistent, indicating an anomaly. In our current implementation, any data labeled as anomalous triggers an alert for the operator.\footnote{Other response measures are possible (although not incorporated), for instance, ignoring the measurements from the station or sending a command to the physical layer to turn off or reset the station.} 

Our digital twin architecture could be useful for various decision-making tasks. For instance, as discussed in \S\ref{sec:cs_impute}, we can use the trained models to simulate (impute) station measurements. This could be handy if weather data from certain stations/regions is missing or very noisy due to faults. Besides, we can synthetically inject noisy data to see how it affects the performance of agricultural prediction tasks, such as predicting surface temperature. In addition, our fruit surface temperature prediction case study finds a reduction in accuracy due to noisy measurements (see \S\ref{sec:cs_fst}).  Hence, models that do not isolate the inconsistent data may trigger false signals to the growers regarding when to take proper cooling precautions to protect their fruits. Failure to detect and isolate noisy sensors (and stations) could burden the growers who rely on the data. A tool like \fancyname that detects inconsistent measurements could be useful for such cases.

\section{Implementation} \label{sec:impl}

The following section presents the current implementation of \fancyname. Our implementation is publicly available: {\tt\url{https://anonymous.4open.science/r/ag-dt-C7BC}}.

%\subsection{System Integration}

\subsection{Physical Layer Integration and Real-Time Data Collection} We integrate \fancyname with a commercial apple orchard located in Quincy, Washington, United States ($47^\circ13^\prime31.3^{\prime\prime}$  $119^\circ57^\prime38.0^{\prime\prime}\text{W}$). The orchard has an average elevation of 322~m and an average slope of $2^\circ$ facing east. The orchard's weather stations are maintained by the Washington State University AgWeatherNet~\cite{awn} team.\footnote{\underline{Note}: The Majority of commercial agricultural weather networks are proprietary. Access to orchards/weather data is non-trivial, and few are publicly accessible for research.} The station where we have access is installed at 1.5~m above ground level. The station uses an all-in-one weather sensor (ATMOS 41 from METER Group~\cite{metergroupATMOSMETER}). It measures several weather parameters such as solar radiation, precipitation, vapor pressure, relative humidity, barometric pressure, horizontal wind speed, wind gust, wind direction, tilt, lightning strike, and lightning average distance.

Weather stations sample measurements at 5-minute intervals and send the data to a vendor cloud service (ZENTRA Cloud~\cite{metergroupZENTRACloud}).  We obtained weather data from March 2023 to January 2025. The collection frequency is twelve data tuples per hour per station (recall: stations send data to the cloud in 5-minute intervals), resulting in 192,613 data points. Following the naming convention suggested by the AgWeatherNet team, we refer to the measurements obtained from the stations as the ``Quincy'' network data. Our physical layer implementation demonstrates the feasibility of integrating \fancyname with a live operational weather network.

\subsubsection{Preexisting Dataset} 

In addition to the live feed obtained from the orchard (Quincy network), we also use a publicly available dataset (named ``Beutenberg'')~\cite{jena_weather_dataset}. This weather data was recorded at the weather stations of the Max Planck Institute for Biogeochemistry in Jena, Germany. The dataset contains 20 different features, such as air temperature, atmospheric pressure, and humidity, between December 2015 to December 2020. The sampling rate for this dataset is six data points per hour. The total number of data points is 266,609. The reason for selecting this dataset is to test our inconsistency classifiers for diverse weather conditions (North America and Europe). Table~\ref{tab:weather} in Appendix~\ref{sec:data_atributes} lists all the attributes obtained from Quincy and Beutenberg weather networks.

\subsection{Implementation of the Digital Twin Layer}

We deploy the digital twin layer on NVIDIA Jetson Orin Developer Kit (12-core Arm Cortex-A78AE 64-bit CPU, 64 GB LPDDR5 RAM, and 2048-core GPU with 64 Tensor Cores)~\cite{karumbunathan2022nvidia}.  The algorithms are implemented in Python and executed on Linux (kernel version 5.15.148). We crawl Quincy weather station data from ZENTRA Cloud through HTTP requests using the Python \ttt{requests} package. The Beutenberg dataset is available as a \ttt{.csv} format, which \fancyname parses and stores locally in its internal database. We use TensorFlow~\cite{abadi2016tensorflow} for model training and inference purposes.

\subsubsection{Inconsistency Generation} \label{sec:inconsistency_generation}

As mentioned in \S\ref{sec:noise}, we use four noise generators to represent imperfect measurements. The formal descriptions of the four noise classes are presented below.

\ul{Random Inconsistency}: The random inconsistency, as the name suggests, represents unpredictable disturbances in sensor measurements with varying intensities. We use random inconsistencies by scaling the weather station measurements through a multiplicative noise. In particular, weather data that have random inconsistencies are denoted as follows:
    % \begin{equation}
      $\mathbf{X}_{\text{random}}(t) = \mathbf{X}(t) \times (1 + \eta_d)$,  
    % \end{equation}
    where \( \eta_d \) is drawn independently from a discrete distribution that contains a density parameter $d$ representing the probability of inconsistency occurrence at each measurement point $t$. For instance, $d = 0.25$ implies that random noise affects one-fourth of the measurements. In our current evaluation (see \S\ref{sec:eval}), \( \eta_d \) is chosen from $[-1.5, 1.5]$. Figure~\ref{fig:faults}(a) depicts a case of 11 random inconsistencies for 1000+ temperature readings.

    \ul{Malfunction Inconsistency}: The readings across multiple sensor measurements can exhibit abnormal fluctuations from their original (expected) values due to errors or disturbances. We recreate this inconsistency using adaptive noise scaling and temporal evaluation, as given in the following equation:
%    \begin{equation}
%         \mathbf{X_{\text{malfunction}}(t)} = \mathbf{X}(t) + \epsilon, \quad \epsilon \sim \mathcal{N}(0, \sigma^2) \times intensity,
%    \end{equation}
	%\begin{equation}
         $\mathbf{X}_{\text{malfunction}}(t) = \mathbf{X}(t) + \epsilon$, 
    % \end{%equation}
    where $\epsilon = \mathcal{N}(0, \sigma^2) \times intensity$ represents the added perturbation drawn from a Gaussian distribution $\mathcal{N}(0, \sigma^2)$ with zero mean and $\sigma^2$ variance. The parameter $intensity$ is a scaling factor used to adjust the magnitude of the noise. In our setup, we consider the intensity factor to be 4.5. Figure~\ref{fig:faults}(b) illustrates this case.

    \ul{Drift Inconsistency}: There may be cases where weather station sensors show gradual, random variations over time. We refer to such variations as drift inconsistencies. Unlike random and malfunction inconsistencies, it follows a pattern, which could be linear or non-linear. The drift inconsistency is mathematically represented as:
    %\begin{equation}
        $\mathbf{X}_{\text{drift}(t)} = \mathbf{X}(t) + \delta + \epsilon$,     
    %\end{equation}    
    where \( \delta \) is an offset (drift intensity), calculated as a multiple of the signal's initial value: $\delta = X(t_0) \times \text{intensity}$. The intensity is selected randomly from a predefined range, e.g., $[-4, 4]$. The last parameter, $\epsilon \sim \mathcal{N}(1, 3\sigma) \times \text{noise\_intensity}$, is an additive Gaussian noise, with $\sigma^2$ being the variance of the signal segment. A visual representation of drift inconsistency is shown in Fig.~\ref{fig:faults}(c).

   \ul{Bias Inconsistency}: This kind of noise shows a consistent deviation in the sensor readings. Bias inconsistency does not vary dynamically and remains constant over time. The mathematical representation of the bias inconsistency is given by:
% \begin{equation}
$\mathbf{X}_{\text{bias}}(t) = \mu \cdot \alpha$,
% \end{equation}
    where $\mu_j = \frac{1}{t_e - t_s} \sum_{t = t_s}^{t_e - 1} \mathbf{X}^{(j)}(t)$ is the mean value of feature $j$ over the selected fault window and $[t_s, t_e)$ is the time interval over which the fault is applied. The parameter $\alpha \in \mathbb{R}$ is the bias intensity factor, which is set to 2 in our implementation. Bias inconsistency is depicted in Fig.~\ref{fig:faults}(d).

\begin{table}[!t]
\caption{Configuration of neural network architectures.}
% \scriptsize
\small
\begin{tabular}{l|l|l|l}  
\toprule
\textbf{Models} & \textbf{Model Layers} & \textbf{Feature Extraction} & \textbf{Dense Layers} \\
%\hline
\midrule
TCN & 8 TCN Blocks & Dilated Convolution & 64, 32, 5 \\
ResNet & 6 Residual Blocks & Conv1D & 256, 5 \\
\hline
LSTM & 2 LSTM Blocks & Sequential & 64, 32, 5 \\
Bi-LSTM & 2 Bi-LSTM Blocks & Bidirectional Sequential & 64, 32, 5 \\
GRU & 2 GRU Blocks & Sequential & 64, 32, 5 \\
\hline
TST & 4 Transformer Blocks & Multi-head Self Attention & 128, 5 \\
Informer & 4 Informer Blocks & ProbSparse Self Attention & 128, 5 \\
\hline
TST-AE & 8 Transformer Blocks & Multi-head Self Attention & 128, 64, 5 \\
LSTM-AE & 6 LSTM Blocks & Sequential & 128, 64, 5 \\
\bottomrule
\end{tabular}
\label{tab:architecture-comparison}
\end{table}

\subsubsection{Learning Model Configurations}

We reconfigure the nine machine learning models for our training and inference purposes.  Table~\ref{tab:architecture-comparison} summarizes the configuration of the learning models. 

We configure the TCN model with 8 causal convolution blocks organized into two stacks: the first stack of 4 blocks with 64 filters each and the second stack of 4 blocks with 32 filters each. For ResNet, we apply 6 residual blocks with three filter sizes (64, 128, 256) across convolution layers to ensure progressive feature extraction and optimize model capacity and computational efficiency.

For RNN-based models, we configure the unidirectional LSTM model using two LSTM layers with 64 and 32 units to capture both short-term and long-term dependencies with a dropout of 0.2 for regularization and to prevent over-fitting. We use a similar setup with two bidirectional LSTM layers and two GRU layers for Bi-LSTM and GRU.

For transformer-based models (\ie TST and Informer), we adopt the multi-head self-attention layers. To balance model capacity and validate effectiveness for time series tasks, TST includes four transformer blocks with a model dimension of 128, a feed-forward dimension of 256 for each block, and 8 attention heads with standard self-attention mechanisms. We use position encoding and a dropout rate of 0.1 for each transformer block. The Informer model utilizes a Prob-sparse self-attention mechanism with similar dimensions and attention heads.

The hybrid auto-encoder architectures (TST-AE and LSTM-AE) include encoder-decoder structures to aim for both classification and reconstruction. TST-AE uses 8 transformer blocks for both encoding and decoding (4 layers each) with 8 multi-head attention. Likewise, LSTM-AE uses 6 LSTM structures with 3 layers in both the encoder and decoder.

\subsubsection{Model Training}

We use partial grid search~\cite{bergstra2012random} to find out the optimized training hyper-parameters (see Table~\ref{tab:hyperparameters}). We train all models with a batch size of 64 and apply early stopping to prevent over-fitting. We also use dropout regularization consistently across all the models, ranging from 0.1 to 0.3, depending on the architecture and robust generalization. Additionally, we standardize the weather parameters (\ie features in the learning models) using a standard scaler to normalize them. The parameter labels are converted to one-hot encoded vectors for compatibility with the categorical cross-entropy loss function. All the models are saved as \texttt{.keras} format. %for further scalibilty test which is described in \S\ref{sec:overhead}

% \begin{wraptable}{r}{0.4\textwidth}
\begin{table}[!t]
% \caption{Hyperparameters used for weather data training.}
\centering
\caption{Hyperparameters used for training.}
% \scriptsize
\small
\begin{tabular}{l | l} 
\toprule
\textbf{Hyperparameter} & \textbf{Value} \\
%\multicolumn{2}{c}{\textbf{Hyperparameter}}  \\
%\hline
\midrule
Learning Rate & 0.001 \\
Batch Size & 64 \\
Dropout & [0.1,0.3]\\
Sequence Length & 48 \\
Optimizer & Adam \\
Loss Function& Categorical cross-entropy\\
\bottomrule
\end{tabular}
%}
%\vspace{4pt}
\label{tab:hyperparameters}
\end{table}

\section{Evaluation} \label{sec:eval}

We perform a comprehensive evaluation to test how digital twins can be useful for checking inconsistent weather data and how each of these nine machine learning models performs (\S\ref{sec:detection}). To this end, we performed two concrete case studies to demonstrate how \fancyname is useful for ag-tech vendors and growers for agricultural decision-making tasks (\S\ref{sec:casestudy}). We test the performance of \fancyname on two fronts: \ca the performance of learning models in detecting inconsistencies in the weather feed, and \cb the overheads and scalability of the inference operations at runtime. 

\subsection{Detecting Inconsistencies} \label{sec:detection}

\subsubsection{Metrics}

To test the performance of the learning models, we consider the following popular statistical metrics.

\begin{itemize}[wide=\parindent, topsep=0pt] 
    \item \textit{Mean Absolute Error (MAE)}: MAE measures the average magnitude of errors between predicted values (\(\hat{y}_i\)) and actual values (\(y_i\)) and is defined as:
    % \begin{equation}
    $\text{MAE} = \frac{1}{n} \sum_{i=1}^{n} |y_i - \hat{y}_i|$.
    % \end{equation}

    \item \textit{Root Mean Square Error (RMSE)}: RMSE measures the square root of the average squared differences between predictions and actual values and is represented as follows:
    % \begin{equation} 
    $\text{RMSE} = \sqrt{\frac{1}{n} \sum_{i=1}^{n} (y_i - \hat{y}_i)^2}$. 

    \item \textit{Coefficient of Determination ($R^2$)}: \( R^2 \) measures the proportion of the variance in the actual values  %that is predictable from the model's predictions. 
    and given by:
    % \begin{equation}
    $R^2 = 1 - \frac{\sum_{i=1}^{n} (y_i - \hat{y}_i)^2}{\sum_{i=1}^{n} (y_i - \bar{y})^2}$,
    % \end{equation}
    where \( \bar{y} \) is the mean of the actual values.

    \item \textit{Precision}: Precision indicates the proportion of true positive predictions among all positive predictions and is defined as:
    % \begin{equation}
    $\text{Precision} = \frac{TP}{TP + FP}$, 
    % \end{equation}
    where $TP$ and $FP$ are true and false positives, respectively.

    \item \textit{Recall}: Recall measures the proportion of true positive predictions among all actual positives and is defined as follows:
    %\begin{equation}
    $\text{Recall} = \frac{TP}{TP + FN}$,
    % \end{equation}
    where $FN$ denotes false negatives.

    \item \textit{$F_1$ Score}: This is a measurement of the harmonic mean of Precision and Recall and is given by:
    %\begin{equation}
    $F_1 = 2 \cdot \frac{\text{Precision} \cdot \text{Recall}}{\text{Precision} + \text{Recall}}$.
    %\end{equation}
\end{itemize}

\begin{table*}[!t]
\caption{Detection performance for the Quincy network.}
\scriptsize 
\setlength{\tabcolsep}{4pt} 
\begin{tabular}{cl|ccc|ccc|ccc}
\toprule
& &\multicolumn{9}{c}{\textbf{Percentage of Inconsistent Data}}\\
\cline{3-11}
\multirow{3}{*}{\textbf{Model}} & \multirow{3}{*}{\textbf{Noise Class}} & \multicolumn{3}{c|}{\textbf{5\%}} & \multicolumn{3}{c|}{\textbf{15\%}} & \multicolumn{3}{c}{\textbf{25\%}} \\
\cline{3-11}
& & \textbf{Precision} & \textbf{Recall} & \textbf{F1} & \textbf{Precision} & \textbf{Recall} & \textbf{F1} & \textbf{Precision} & \textbf{Recall} & \textbf{F1} \\
\midrule
\multirow{4}{*}{TCN} & Random & 0.9712 & 0.8441 & 0.9032 &  0.9963 & 0.9890 & 0.9926  & 0.9905 & 0.9949 & 0.9927 \\
& Malfunction & 0.7952 &  0.4946 & 0.6098 & 0.9750 & 0.9292 & 0.9515 & 0.9843 & 0.9837 & 0.9840 \\
& Drift & 0.9887  &  0.9782  &  0.9834 & 0.9979 & 0.9993 & 0.9986 & 0.9979 & 1.0000 & 0.9989  \\
& Bias & 1.0000 & 0.2622 & 0.4154 & 0.9992 & 0.9970 & 0.9981 & 0.9993 & 0.9993 & 0.9993\\
\midrule
\multirow{4}{*}{ResNet} & Random & 0.9722 & 0.9272 & 0.9492 & 0.9971 & 0.9949 & 0.9960 & 0.9891 & 1.0000 & 0.9945 \\
& Malfunction & 0.9214 & 0.6792 & 0.7820 & 0.9734 & 0.9462 & 0.9596  & 0.9685 & 0.9857 & 0.9770  \\
& Drift & 1.0000 & 0.9579 & 0.9785 & 1.0000 & 1.0000 & 1.0000 & 0.9993 & 1.0000 & 0.9996\\
& Bias & 0.9497 & 0.7356 & 0.8290 & 0.9985 & 0.9970 & 0.9978  &  1.0000 & 0.9903 & 0.9951 \\
\midrule
\multirow{4}{*}{LSTM} & Random & 0.9875 & 0.9904 & 0.9890 & 0.9978 & 0.9971 & 0.9974 & 0.9949 & 1.0000 & 0.9974  \\
& Malfunction & 0.9753&  0.4850 & 0.6479 & 0.9843 & 0.9407 & 0.9620 &  0.9848 & 0.9721 & 0.9784 \\
& Drift & 0.9993 & 1.0000 & 0.9996  & 1.0000 & 1.0000 & 1.0000 & 1.0000 & 1.0000 & 1.0000 \\
& Bias & 1.0000 & 0.3828 & 0.5536 & 0.9970 & 1.0000 & 0.9985 &  1.0000 & 0.9985 & 0.9993\\
\midrule
\multirow{4}{*}{Bi-LSTM} & Random & 0.9769 & 0.9316 & 0.9537 & 0.9971 & 0.9978 & 0.9974 &  0.9827 & 1.0000 & 0.9913\\
& Malfunction & 0.9182 & 0.5279 & 0.6704 & 0.9707 & 0.9707 & 0.9707 & 0.9665 & 0.9639 & 0.9652 \\
& Drift & 1.0000 & 0.9593 & 0.9792 & 1.0000 & 1.0000 & 1.0000 & 0.9986 & 1.0000 & 0.9993 \\
& Bias & 0.9976 & 0.9176 & 0.9559 & 1.0000 & 0.9993 & 0.9996 & 1.0000 & 0.9955 & 0.9977 \\
\midrule
\multirow{4}{*}{GRU} & Random & 0.9790 & 0.9603 & 0.9696 &  0.9862 & 0.9978 & 0.9920 & 0.9956 & 1.0000 & 0.9978\\
& Malfunction & 0.9839 & 0.5000 & 0.6631 & 0.9655 & 0.9346 & 0.9498 & 0.9890 & 0.9823 & 0.9856 \\
& Drift & 1.0000 & 0.9993 & 0.9996 & 1.0000 & 1.0000 & 1.0000  & 1.0000 & 0.9993 & 0.9996 \\
& Bias & 0.9985 & 0.9888 & 0.9936 & 0.9926 & 0.9985 & 0.9955 & 1.0000 &  0.9978 & 0.9989 \\
\midrule
\multirow{4}{*}{TST} & Random & 0.8289 & 0.8765 & 0.8520 &  0.9832 & 0.9868 & 0.9850 & 0.9854 & 0.9949 & 0.9901 \\
& Malfunction & 0.8701 & 0.6935 & 0.7718 & 0.9486 & 0.8678 & 0.9064 & 0.9465 & 0.9762 & 0.9611  \\
& Drift &  0.9986 & 0.9712 & 0.9847 & 0.9785 & 0.9909 & 0.9847 & 0.9814 & 0.9972 & 0.9892 \\
& Bias & 0.9945 & 0.9446 & 0.9689  & 1.0000 & 0.9963 & 0.9981 &0.9985 & 0.9963 & 0.9974 \\
\midrule
\multirow{4}{*}{Informer} & Random & 0.9705 & 0.8941 & 0.9307 & 0.9805 & 0.9985 & 0.9894 & 0.9876 & 0.9978 & 0.9927\\
& Malfunction & 0.9898 & 0.9516 & 0.9703 & 0.9753 & 0.9666 & 0.9709 & 0.9015 & 0.9789 & 0.9386 \\
& Drift & 0.9898 & 0.9516 & 0.9703 & 0.9979 & 0.9993 & 0.9986  & 0.9666 & 0.9965 & 0.9813 \\
& Bias & 0.8843 & 0.3551 & 0.5067 & 0.9985 & 0.9978 & 0.9981 & 1.0000 & 1.0000 & 1.0000 \\
\midrule
\multirow{4}{*}{TST-AE} & Random & 0.6393 & 0.6346 & 0.6369 & 0.9350 & 0.9831 &  0.9584 &  0.9414&0.9919&0.9660\\
& Malfunction & 0.3593 & 0.2732 & 0.3104 & 0.9514 & 0.9060 & 0.9281 & 0.9116&0.9135&0.9126\\
& Drift & 0.9986 & 0.9733 & 0.9858 & 0.9965 & 0.9860 & 0.9912  & 0.9746&0.9972&0.9858  \\
& Bias & 0.9888 & 0.3311 & 0.4961 & 0.9901 & 0.9723 & 0.9811 & 0.9977&0.9880&0.9928 \\
\midrule
\multirow{4}{*}{LSTM-AE} & Random & 0.9437 & 0.8007 & 0.8663 & 0.9948 & 0.9926 & 0.9937 & 0.9934 & 0.9963 & 0.9949 \\
& Malfunction & 0.7480 & 0.4550 & 0.5659  & 0.9126 & 0.9816 & 0.9458 & 0.9403 & 0.9864 & 0.9628\\
& Drift & 0.9672 & 0.9923 & 0.9796 &1.0000 & 0.9979 & 0.9989& 0.9965 & 1.0000 & 0.9982 \\
& Bias & 0.9706 & 0.8644 & 0.9144 & 0.9992 & 0.9978 & 0.9985  & 1.0000 & 0.9993 & 0.9996\\
\bottomrule
\end{tabular}
\label{tab:quincy}
\end{table*}
\begin{table*}[!t]%[htbp]
\caption{Detection performance for the Beutenberg network.}
\scriptsize 
\setlength{\tabcolsep}{4pt} 
\begin{tabular}{cl|ccc|ccc|ccc}
\toprule
& &\multicolumn{9}{c}{\textbf{Percentage of Inconsistent Data}}\\
\cline{3-11}
\multirow{3}{*}{\textbf{Model}} & \multirow{3}{*}{\textbf{Noise Class}} & \multicolumn{3}{c|}{\textbf{5\%}} & \multicolumn{3}{c|}{\textbf{15\%}} & \multicolumn{3}{c}{\textbf{25\%}} \\
\cline{3-11}
& & \textbf{Precision} & \textbf{Recall} & \textbf{F1} & \textbf{Precision} & \textbf{Recall} & \textbf{F1} & \textbf{Precision} & \textbf{Recall} & \textbf{F1} \\
\midrule
\multirow{4}{*}{TCN} & Random &  0.9825&0.6961&0.8149  & 0.9922&0.9798&0.9860& 0.9738&0.9969&0.9852 \\
& Malfunction &0.9004&0.8382&0.8682&0.9183&0.9907&0.9532&  0.9635&0.9791&0.9713 \\
& Drift &0.9975&0.9365&0.9660&0.9962&0.9981&0.9972 & 0.9337&1.0000&0.9657 \\
& Bias & 0.9214&0.6065&0.7315 &0.9886&0.9353&0.9612 & 0.9994&0.8620&0.9256  \\
\midrule
\multirow{4}{*}{ResNet} & Random &  0.9448&0.9385&0.9417  & 0.9948&0.9933&0.9941 &0.9964&0.9990&0.9977 \\
& Malfunction &0.9009&0.9235&0.9121 &0.9153&0.9722&0.9429 & 0.9812&0.9903&0.9857 \\
& Drift & 1.0000&0.4612&0.6312 & 0.9991&1.0000&0.9995 &0.9482&1.0000&0.9734\\
& Bias & 0.5690&0.7515&0.6476  & 0.9988&0.9364&0.9666 &  0.9994&0.9342&0.9657 \\
\midrule
\multirow{4}{*}{LSTM} & Random & 0.9884&0.9669&0.9775 & 0.9995&0.9969&0.9982 &  0.9969&0.9979&0.9974 \\
& Malfunction & 0.9642&0.8479&0.9023 & 0.9310&0.9954&0.9621 & 0.9289&0.9940&0.9604 \\
& Drift & 0.9591&0.9920&0.9752 & 1.0000&1.0000&1.0000 & 0.9981&1.0000&0.9991\\
& Bias &0.9925&0.9272&0.9588& 0.9994&0.9364&0.9669 & 0.9994&0.9342&0.9657 \\
\midrule
\multirow{4}{*}{Bi-LSTM} & Random & 0.9851&0.9922&0.9887 &0.9882&0.9959&0.9920  &  0.9979&0.9984&0.9982 \\
& Malfunction & 0.9302&0.8275&0.8759 &0.9208&0.9917&0.9549 & 0.8052&0.9986&0.8916\\
& Drift &1.0000&0.9878&0.9938&  0.9991&1.0000&0.9995& 1.0000&0.9995&0.9998 \\
& Bias & 0.9954&0.9288&0.9610 & 1.0000&0.9272&0.9622 &  0.9994&0.9348&0.9660\\
\midrule
\multirow{4}{*}{GRU} & Random & 0.9684&0.9974&0.9827 & 0.9974&0.9969&0.9972 & 0.9954&0.9984&0.9969 \\
& Malfunction &0.9502&0.8493&0.8969 & 0.9384&0.9745&0.9561 &  0.9096&0.9986&0.9520 \\
& Drift & 0.9995&0.9864&0.9929 &  1.0000&1.0000&1.0000 & 1.0000&1.0000&1.0000 \\
& Bias &0.9852&0.9326&0.9582 & 0.9983&0.9364&0.9663 & 1.0000&0.9321&0.9648 \\
\midrule
\multirow{4}{*}{TST} & Random & 0.9875&0.8956&0.9393 & 0.9734&0.9850&0.9792 & 0.9046&0.9902&0.9455 \\
& Malfunction & 0.7639&0.8340&0.7974 & 0.9344&0.9504&0.9423 & 0.8996&0.9764&0.9364 \\
& Drift & 0.9804&0.6367&0.7720 &0.9624&1.0000&0.9808 & 0.9851&0.9958&0.9904 \\
& Bias & 0.5825&0.5881&0.5853 & 0.9977&0.9353&0.9655 & 0.9954&0.9278&0.9604 \\
\midrule
\multirow{4}{*}{Informer} & Random & 0.9731&0.9147&0.9430 & 0.9595&0.9922&0.9756& 0.9660&0.9835&0.9746 \\
& Malfunction & 0.9392&0.7955&0.8614& 0.9281&0.9870&0.9566 & 0.9132&0.9856&0.9480 \\
& Drift &0.9482&0.9995&0.9732&0.9668&0.9995&0.9829& 0.9449&0.9995&0.9157 \\
& Bias & 0.9960&0.7973&0.8856 &0.9965&0.9100&0.9513 & 0.9987&0.9005&0.9887 \\
\midrule
\multirow{4}{*}{TST-AE} & Random & 0.9500&0.8346&0.8886 & 0.8934&0.9576&0.9244  & 0.9127&0.9938&0.9515 \\
& Malfunction & 0.9075&0.5684&0.6990& 0.8821&0.8878&0.8849 &0.9253&0.9875&0.9554 \\
& Drift & 0.9295&0.9365&0.9330 & 0.9312&0.9939&0.9615 &  0.9939&1.0000&0.9970\\
& Bias & 0.8945&0.8410&0.8669 & 0.9821&0.9181&0.9490 &1.0000&0.9235&0.9602\\
\midrule
\multirow{4}{*}{LSTM-AE} & Random &0.9216&0.9602&0.9405&0.9857&0.9943&0.9900& 0.9984&0.9959&0.9972 \\
& Malfunction &0.8356&0.8391&0.8374& 0.9587&0.9889&0.9735 &  0.9764&0.9972&0.9867 \\
& Drift &0.9924&0.8598&0.9213& 0.9645&0.9972&0.9806 &  0.9478&1.0000&0.9732 \\
& Bias &0.9139&0.8755&0.8943 & 1.0000&0.9278&0.9625 & 0.9994&0.9358&0.9666 \\
\bottomrule
\end{tabular}
\label{tab:beutenburg}
\end{table*}

\subsubsection{Findings}

We evaluate the performance of the neural network models with a varied amount of noise using the metrics introduced above. We vary the percentage of injected inconsistent data (5\%, 15\%, and 25\%) in the training samples across all four classes (random, malfunction, drift, and bias). Then we observe whether each of the nine models can detect inconsistent measurements. All the classes have an equal percentage of inconsistency (\ie each class roughly contributed to one-fourth of the inconsistent samples). For a given observation window, about 15\% of testing (inference) samples have inconsistent readings. Table~\ref{tab:quincy} and Table~~\ref{tab:beutenburg} demonstrate the comparative performance of the learning models for Quincy and Buetenburg networks, respectively.

% \mhtodo{Waiting for Tamim's clarification on experiment setup.}

The performance of the models is influenced by the type of inconsistency class. For all the models, bias inconsistency is the easiest to detect, as these kinds of inconsistencies show a shifting pattern (scaled offset) from expected values. We find that for bias inconsistencies, all models achieve $F_1$ scores of 1 (or close to 1) for all three inconsistency levels (\ie 5\% to 25\%). For lower noisy samples (\eg 5\%) the models often perform poorly. When there is smaller noisy data in the training samples, those models struggle to learn. Most models achieve good performance in terms of precision, recall, and $F_1$ score when there are enough noisy samples (\eg 15\% and 25\% inconsistency levels). In contrast to bias, malfunction inconsistency shows erratic behavior and is challenging to identify. For example, the recall of TST-AE drops to 0.2732 and 0.5684 at 5\% for Qunicy and Beutenburg networks, respectively. Several models face difficulties in detecting this fault as it lacks a clear pattern, particularly at lower inconsistency levels. When the percentage of noisy samples increased in the training, some models showed decent performance. For instance, ResNet achieves a precision of 0.9685 and 0.9812 at 25\% for both networks. Random noise does not have predictable patterns, and a higher number of faulty samples aids in detection performance. For the remaining inconsistency classes (\ie drift), the classifiers perform relatively well, as when inconsistencies are triggered, they show temporal patterns. ResNet and LSTM show more consistent performance (fewer variations) for all inconsistency types, but other models sometimes have high precision but may have lower recall and $F_1$ scores. We believe this is due to the internal architectures of the models, which cause more false negatives. However, all models show consistent performance for both weather networks.

\begin{tcolorbox}[colback=gray!10, colframe=black, boxrule=0.3pt,left=2.5pt,right=2.6pt,top=1.5pt,bottom=1.5pt]
Increasing faulty training samples improved the detection of
inconsistencies. If we have large data for training the models, all models work fairly well, although ResNet and LSTM show the most consistent performance.
\end{tcolorbox}

\begin{figure}[!t]
\centering
\includegraphics[scale=0.7]{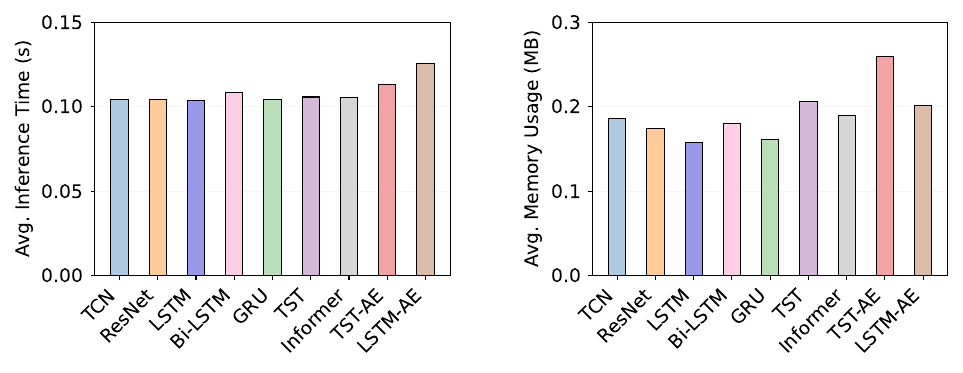}
% \vspace{5pt}
\caption{Average inference time (left) and average memory usage  (right) per sample for the Quincy network.}
\label{fig:overhead_no_container}
\end{figure}

\subsection{Inference Overheads and Scalability}\label{sec:overhead}

We also measure the overhead of inferring inconsistencies in weather measurements. 
%Specifically, we measured inference time and memory overheads 
%\subsubsection{Inference Time}
As \fancyname is designed to check real-time inconsistencies, it is crucial to know how quickly we can identify anomalous data and how much memory resource is needed. Hence, we measure the runtime inference overheads (inference time and memory consumption) on our development platform (Jetson Orin). We use Python \verb|time.time()| function to obtain inference times. We observe one sequence length (48 samples), repeat this 100 times, and calculate the average to get the inference times. 
%The overall range of inference time is slightly different for the two network measurements due to variations in the number of attributes and size of the datasets. 
As the left of Fig.~\ref{fig:overhead_no_container} shows, we can infer inconsistent data in less than 10 ms, even in moderately powerful off-the-shelf hardware. The right plot of Fig.~\ref{fig:overhead_no_container} shows the average memory consumption per sample obtained by Python \verb|memory-profiler| tool. The numbers were obtained from 5,000 samples and an average of 10 runs. Our numbers suggest ANN models (ResNet, LSTM and GRU) are the most memory-efficient. Hybrid models are more memory-intensive due to their encoder-decoder structures. However, average memory consumption is less than 0.4 MB, which is insignificant considering the memory availability in modern hardware.

\begin{figure}[!t]
\centering
\includegraphics[scale=0.7]{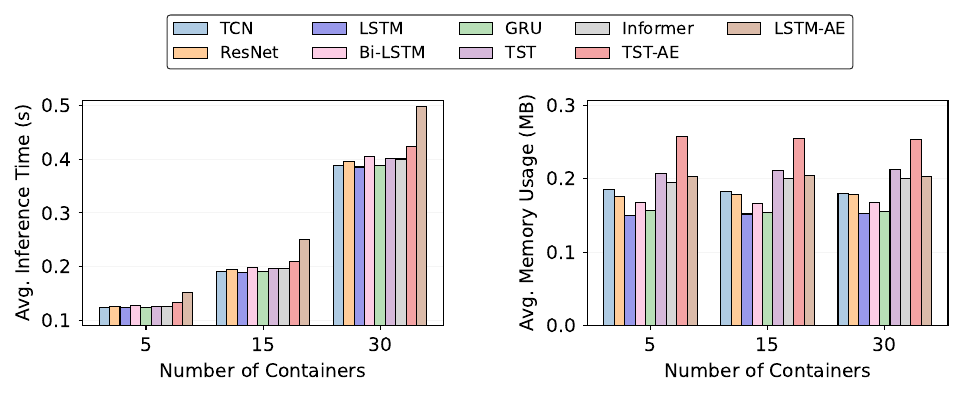}
% \vspace{5pt}
\caption{Performance of \fancyname with multiple instances. The plots show average inference time (left) and average memory usage  (right) for the Quincy network with varying container configurations. Even with 30 instances of \fancyname running concurrently, we can get inference results in less than a second and no more than 0.4 MB of memory consumption.}
\label{fig:overhead_container}
\end{figure}

In the next set of experiments (Fig.~\ref{fig:overhead_container}), we study the scalability of \fancyname. Specifically, we want to test if \fancyname can be used as a ``digital twin as-a-service'' model and serve various ag-tech vendors simultaneously. While the timing and memory overheads can be minimized by using powerful server-grade machines, we want to evaluate the effectiveness of our digital twin design in a relatively low-configured, low-cost device (such as our development platform, Jetson Orin, which has moderate CPU and GPU power and costs about \$2000). To test scalability, we create multiple instances of \fancyname using Docker containers~\cite{turnbull2014docker}. Each container instance is a representation of the \fancyname digital twin ``service'' provided to the end-users. We deploy three configurations of  $\{5, 15, 30\}$ containers and instantiate them using Docker Compose (version 1.29.2). Each container image is identical and configured with 1 CPU core and 4~GB of memory.

As expected, when multiple instances are running simultaneously, the inference time increases (left plot in Fig.~\ref{fig:overhead_container}). This is because, ultimately, the inference processes are scheduled by the Linux kernel. When there are more processes, the system load increases, which leads to higher inference times. However, as shown in the right plot of Fig.~\ref{fig:overhead_container}, memory consumption is relatively unchanged with increasing number of services. This is because we load a fixed batch size of input samples for inference, and for each container, the data loaded into memory remains the same. Besides, as we observe for the single instance case (Fig.~\ref{fig:overhead_no_container}), memory usage is relatively low (less than 0.4 MB) for all the models. Hence, even with more services---factoring in the memory overheads of Docker itself---it does not stress the Jetson. Our findings suggest that it is feasible to deploy multiple instances of \fancyname with ease, even in a relatively low-configuration, non-server-grade machine like Jetson.

\begin{tcolorbox}[colback=gray!10, colframe=black, boxrule=0.3pt,left=2.5pt,right=2.6pt,top=1.5pt,bottom=1.5pt]
    We can execute multiple instances of \fancyname without significant overhead. In particular, on the Jetson board, \ca the query time for inferring inconsistent measurements in our test networks is less than a second, and \cb the memory usage is less than 0.4 MB when 30 instances of \fancyname are running concurrently on a containerized Linux environment.
\end{tcolorbox}

\section{Using \fancyname for Agricultural Decision-Making: Case Studies} \label{sec:casestudy}

The goal of this research is to demonstrate how a digital twin architecture can be useful for agricultural decision-making tasks when perfect environmental information is not available. We now demonstrate the usefulness of \fancyname using two agriculture-specific case studies.

\begin{figure}[!t]
\centering
\includegraphics[scale=0.145]{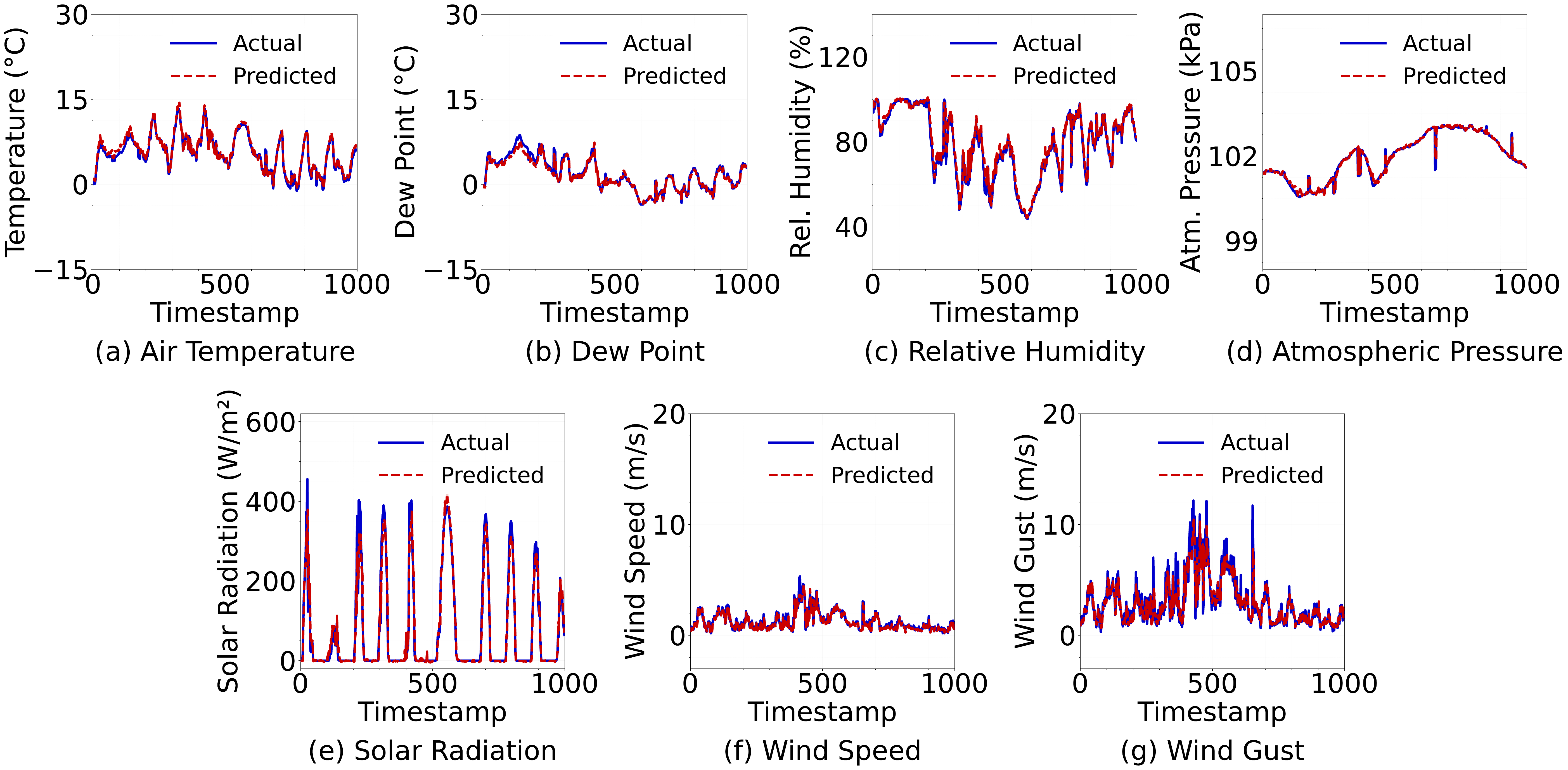}
\caption{Predicting weather attributes using a generative model.}
\label{fig:prerdiction}
\end{figure}

% \begin{wraptable}{r}{0.41\textwidth}
\begin{table}[!t]
%\caption{Error statistics of weather data prediction.}
\caption{Prediction error statistics.}
\centering
% \scriptsize
\small
\begin{tabular}{l | c | c | c} 
\toprule
\textbf{Attribute} & \textbf{MAE} & \textbf{RMSE} & \textbf{\textit{R}\textsuperscript{2}} \\
\midrule
Air Temperature & 0.5446 & 0.7131 & 0.9804 \\
Dew Point & 0.5805 & 0.8150 & 0.9735 \\
Atmospheric Pressure & 0.0919 & 0.1324 & 0.9765 \\
Relative Humidity & 1.4010 & 2.1400 & 0.9756 \\
Solar Radiation & 9.2337 & 20.3374 & 0.9564 \\
Wind Speed & 0.1881 & 0.2700 & 0.8065 \\
Gust Speed & 0.4573 & 0.6466 & 0.7767 \\
\bottomrule
\end{tabular}
\label{tab:errormetrics}
\end{table}
%\end{wraptable}

\subsection{Case Study 1: Weather Data Imputation} \label{sec:cs_impute}

Often, weather data becomes unavailable due to faults or breaches in the sensors or the entire station~\cite{boomgard2022machine}. In addition, often, the end users may not have enough regional or site-specific data to use for analytics. In such cases, it is useful to predict sensor readings that can be ``imputed'' for missing or inconsistent results.  

Our first case study shows the use of \fancyname to predict future weather measurements. Depending on how the models are trained, \fancyname can also mimic certain inconsistency patterns in sensor measurements. To predict sensor data for imputation, we use a generative model (C-RNN-GAN~\cite{mogren2016c}), as this model is better suited for capturing short-term and long-term dependencies in time-series data. 
% LSTM is used as generator which takes in sequences of historical weather data and generates prediction for the next time step. The discriminator utilizes a Bi-LSTM network to distinguish between real and predicted data to evaluate sequences with a time dimension of one to validate authenticity of the generator's output. We have weighted the prediction loss ten times more than the adversarial loss to optimize predicting accuracy.
The blue lines in Fig.~\ref{fig:prerdiction} show the ground truth data from the orchard (Quincy network), and the red dotted lines represent our predicted values obtained from the generative model. As shown in Fig.~\ref{fig:prerdiction} and based on the error metrics presented in Table~\ref{tab:errormetrics}, the prediction results are pretty accurate.

% Figure~\ref{fig:pipeline}) \hl{flags any inconsistency in real-time data, the window will be replaced by the generator of the GAN model and send the to historical database or any other domain-specific module}(e.g., minimizing error in apple surface fruit temperature in \S\ref{sec:cs_fst}). This case study further demonstrates how users can leverage \fancyname to predict site-specific weather that can be fed to other decision-making tasks (for example, surface temperature prediction) when measurements are not available for a target orchard/farm. 

\begin{figure}[!t]
\centering
\includegraphics[scale=0.77]{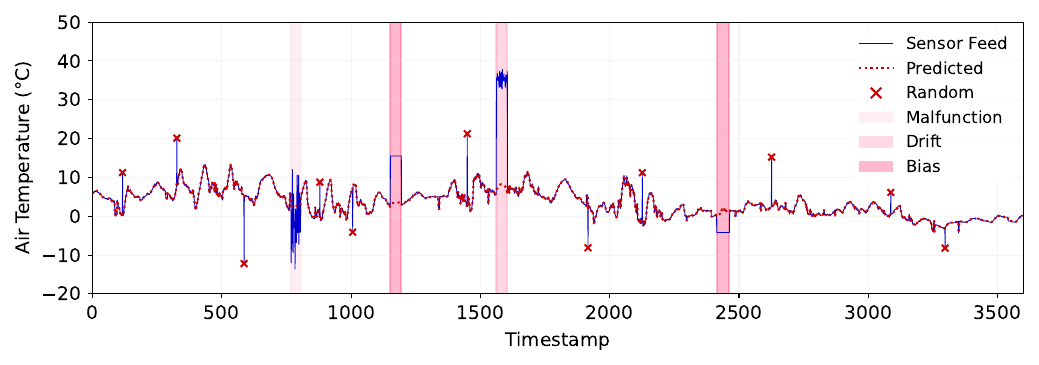}
\caption{Imputing a weather attribute (air temperature) when real-time sensor feeds are imperfect. We assume weather sensors produce inconsistent data at any random point in time (denoted by red shaded regions).}
\label{fig:imputation}
\end{figure}

We also conduct experiments to demonstrate the use of data imputation when the station data is imperfect (see Fig.~\ref{fig:imputation}). We use seven weather attributes (air temperature, dew point, humidity, atmospheric pressure, solar radiation, wind speed, and wind gust) available from the Quincy network. For this experiment, we inject inconsistent samples at different times of the observation window (the red shaded regions) and analyze whether \fancyname can detect faulty samples and impute expected measurements. The blue lines in the figure show real-time weather data feeds. When the consistency checker (Block~\circled{2} in Fig.~\ref{fig:pipeline}) flags any real-time weather feed as inconsistent, that data is replaced by the imputed data (\eg the red dotted lines in the figure).  This case study demonstrates how users can leverage \fancyname to predict site-specific weather that can be fed to other decision-making tasks %(for example, surface temperature prediction) 
when measurements are not available for a target orchard/farm. As a concrete example, we also show how imputed measurements are useful for decision-making tasks, for instance, minimizing error in fruit surface temperature prediction, as we discuss next in \S\ref{sec:cs_fst}. %\mhtodo{Not done yet}

\subsection{Case Study 2: Fruit Surface Temperature Prediction} \label{sec:cs_fst}

Heat stress to maturing fruit is a key concern for tree fruit growers. Localized weather-based fruit surface temperature (an indicator of fruit stress) can help in planning better mitigation strategies and ultimately reduce crop losses. To reduce fruit stress, growers use overhead cyclic rotating sprinklers, foggers, netting, and protectant sprays~\cite{racsko2012sunburn}. However, precise data inputs are needed for the actuation and effective use of some of these mitigation techniques. Faulty or incorrect weather estimations can affect the initiation of such protective measures. We now show how \fancyname could be useful to predict fruit surface temperature so that growers can make informed decisions.

We use the aforementioned nine learning models to predict the apple surface
temperatures from the weather measurements obtained from the Quincy network. The weather parameters for surface temperature prediction include canopy air temperature ($^\circ$C), wind speed (m/s), dew point ($^\circ$C), and solar radiation (W/m\textsuperscript{2}). Past research shows fruit surface temperature can be estimated from these weather attributes~\cite{li2014modeling}. Thanks to the AgWeatherNet team, we also obtained the ``ground truth'' apple temperature measurements. 
%canopy air temperature ($^\circ$C), wind speed (m/s), dew point ($^\circ$C), and solar radiation (W/m\textsuperscript{2}). 
We study for both perfect and imperfect weather feeds. For imperfect cases,
approximately 20\% of the weather sensor feeds are faulty/inconsistent samples, and like before, we use equal noise splits (\ie 5\% for each class). 
%\mhtodo{Need to fix that with integers or equal split. May need new experiments}

\begin{table*}[!t]
\scriptsize
\caption{Using \fancyname to predict apple surface temperature.} 
%With, Without and Replacing Inconsistent Measurements}
\begin{tabular}{l | C{3em} C{3em} C{3em} | C{3em} C{3em} C{3em} | C{3em} C{3em} C{3em}}
\toprule
& \multicolumn{3}{c|}{\textbf{No Imperfection}} & \multicolumn{3}{c|}{\textbf{Imperfect Measurements}} & \multicolumn{3}{c}{\textbf{Imputing Inconsistencies}} \\
\midrule
\textbf{Models} & \textbf{MAE} & \textbf{RMSE} & \textbf{\textit{R}\textsuperscript{2}} & \textbf{MAE} & \textbf{RMSE} & \textbf{\textit{R}\textsuperscript{2}} & \textbf{MAE} & \textbf{RMSE} & \textbf{\textit{R}\textsuperscript{2}} \\
\midrule
TCN & 0.6874 & 1.2911 & 0.9288 & 1.9347 & 4.9019 & 0.2634 & 0.7652 & 1.3466 & 0.9226 \\

ResNet & 0.5823 & 1.2395 & 0.9344 & 2.1711 & 5.4140 & 0.1014 & 0.6695 & 1.3038 & 0.9274\\
\midrule
LSTM & 0.8335 & 1.3969 & 0.9167 & 1.9208 & 4.1776 & 0.4650 & 0.9215 & 1.4688 & 0.9079 \\

Bi-LSTM & 1.0283 & 1.5890 & 0.8922 & 1.9139 & 3.8689 & 0.5411 & 1.0959 & 1.6467 & 0.8842 \\

GRU & 1.7242 & 2.1495 & 0.8027 & 2.3225 & 3.8069 & 0.5557 & 1.8041 & 2.2292 & 0.7879 \\
\midrule
TST & 0.8128 & 1.3729 & 0.9195 & 2.4770 & 5.3600 & 0.1193 & 0.9348 & 1.8015 & 0.8615 \\

Informer & 0.5983 & 1.2530 & 0.9330 & 2.2968 & 5.3474 & 0.1234 & 0.6769 & 1.3122 & 0.9265\\
\midrule
TST-AE & 1.5949 & 1.9784 & 0.8329 & 2.3072 & 3.5205 & 0.6201 & 1.6834 & 2.0721 & 0.8167 \\

LSTM-AE & 0.9393 & 1.4239 & 0.9134 & 1.7415 & 3.4767 & 0.6295 & 1.0146 & 1.4949 & 0.9046 \\
\bottomrule
\end{tabular}
\label{tab:fst}
\end{table*}

In the absence of faulty data (\ie column labeled with ``No Imperfection'' in Table~\ref{tab:fst}), CNN and Transformer-based models (in particular, ResNet and Informer) show best performance with the lowest MAE, RMSE, and $R^2$ numbers. Based on these findings, we conclude that convolutional and attention-based models perform well on clean (consistent) weather data. We then check for cases where we have noisy readings (the ``Imperfect Measurement'' column in Table~\ref{tab:fst}). As Table~\ref{tab:fst} shows, the performance of the models
(MAE, RMSE, and $R^2$) drops in the presence of inconsistent
measurements. If we perform predictions ignoring faulty measurements, it can lead to incorrect decision-making for farmers who rely on the apple surface temperature values to initiate proper protective actions (such as adjustment of irrigation or covering fruits). Hence, we also test how a digital twin can assist in this case. The last column in Table~\ref{tab:fst} (``Imputing Inconsistencies'') shows the performance numbers when \fancyname replaces faulty measurements with imputed values (for instance, using a generative model, as presented in \S\ref{sec:cs_impute}). As the table shows, imputing faulty samples with the expected value improves overall prediction performance (\ie the errors are lower than in the imperfect case). For this particular prediction problem, our experiments show  GRU, TST, and TST-AE have more variability (higher MAE and RMSE but lower $R^2$). We attribute this to the model's internal architectures, which cause them to learn less efficiently (larger errors). Overall, ResNet and Informer models show better results for predicting the apple fruit surface temperature in noisy measurements. This case study further shows how growers can leverage the modularity of \fancyname and customize it for weather-driven decision-making problems.

\section{Related Work}

 Automating the modern agriculture sector is a promising research area. Zhai~\etal~\cite{zhai2020decision} present a review of agricultural decision support systems by incorporating IoT, artificial intelligence, and big data. The use of digital twins for the agriculture sector is also studied. Peladarinos~\etal~\cite{peladarinos2023enhancing} explore the role of digital twin in smart farming and emphasize their ability to create virtual replicas of agricultural systems by integrating sensors and remote imagery. The authors show that digital twins can support decision-making in irrigation, fertilization, and pest control using dynamic visualizations. Verdouw~\etal~\cite{verdouw2021digital} propose a framework to integrate general systems control with IoT-based architecture to manage agricultural farm operations dynamically, which can enhance productivity, sustainability, and decision-making processes. An IoT-enabled digital twin pipeline is proposed~\cite{alves2019digital} that study integrates environmental data collection, cloud-based data processing, and real-time visualization to enhance resource management and optimize farming processes. None of the existing research focuses on investigating data anomalies in agricultural weather networks.

Digital twin technology is used in many cyber-physical domains for anomaly detection~\cite{castellani2020real,gao2021anomaly,xu2023digital}. Castellani~\etal~\cite{castellani2020real} propose a weakly-supervised anomaly detection technique for industrial systems using digital twins. The digital twin is used to generate synthetic data that represents normal operation conditions. The authors then use a set of labeled anomalies from real data to enhance detection accuracy. In a similar direction, Gao~\etal~\cite{gao2021anomaly} introduce an anomaly detection framework for digital twin-driven cyber-physical systems, especially those that are vulnerable to sensor faults and model discrepancies. By integrating simulation data (from the digital twin) with real-time sensor data (from the physical plant), the proposed approach enables the detection of anomalies due to both sensor issues and model inaccuracies. LATTICE~\cite{xu2023digital} is a digital twin-based anomaly detection framework that uses curriculum learning to improve anomaly detection in cyber-physical systems. However, the above-mentioned techniques do not consider agricultural use cases. To the best of our knowledge, \fancyname is one of the first explorations for resiliency analysis of weather-driven agricultural systems.

\section{Conclusion}

%This research introduces 
We introduce a novel use of digital twin technology to analyze data inconsistency issues in agricultural weather networks, which is a key attribute for many decision-support models and smart automation systems. 
%The proposed digital twin of agricultural weather networks---\fancyname---
\fancyname will be a fundamental tool for twinning physical entities
(weather stations) to provide in silico emulation capabilities and deliver insights to farm decision-making. By addressing the challenges of data inconsistency, \fancyname establishes a foundation for integrating digital technologies into agriculture. We envision that a framework like \fancyname will have far-reaching insights into many cyber-agriculture aspects, such as data-driven analytics and resiliency, cybersecurity, and vulnerability analysis.

\begin{comment}
In this paper, we propose a digital twin framework for real-time anomaly detection, fruit surface temperature prediction in agricultural weather networks and run time behavior of different deep learning models in terms of inference time and memory usage. The framework integrates real-time data with synthetic data through TCN-based simulation. This study highlights the adaptability of the our framework to different fault types and measured its performance on two different dataset to validate the reliability of it. Our digital twin approach is capable of classifying sensor faults with high precision, recall and F1 score and predict FST in certain fault conditions. The measure of the inference time and memory usage also give insights which model is suitable for resource aware environments. Overall, this study contributes to the advancement of digital twin applications in agriculture with improved data consistency, fault management, predictive modeling and decision-making in environmental monitoring systems.
\end{comment}

% \section*{Acknowledgment}

% The authors thank the AgWeatherNet team for providing access to the Quincy orchard weather stations and introducing them to this crucial agricultural domain-specific problem.

\begin{acks}
We thank WSU AgWeatherNet team (Lav Khot and Srikanth Gorthi) for sharing weather station data. This research is partly supported by the U.S. National Science Foundation Award 2345653. Any findings, opinions, recommendations, or conclusions expressed in the paper are those of the authors and do not necessarily reflect the sponsor’s views.
\end{acks}

%\newpage
\bibliographystyle{ieeetr}
\bibliography{references/ref_mh,references/reference}

\appendix
\newpage
\section{Appendix}

\subsection{Dataset Attributes} \label{sec:data_atributes}

Table~\ref{tab:weather} lists the weather data attributes from Quincy and Beutenberg weather stations.
\vspace{-3pt}

\begin{table}[!h]
\caption{Weather Data Attributes}
\footnotesize
\label{tab:weather}
\begin{tabular}{|p{5cm}|p{6.25cm}|}
\hline
\multicolumn{1}{|c|}{\textbf{Quincy}} & \multicolumn{1}{c|}{\textbf{Beutenberg}} \\ 
\hline\hline
Air Temperature ($^{\circ}$C) & Air Temperature ($^{\circ}$C) \\ 
Atmospheric Pressure (kPa) & Potential Temperature (K) \\ 
Vapor Pressure (kPa) & Dew Point ($^{\circ}$C) \\ 
Dew Point ($^{\circ}$C) & Logger Temperature ($^{\circ}$C)\\
Vapor Pressure Deficit (kPa) & Vapor Pressure (mbar) \\ 
Reference Pressure (kPa) & Maximum Vapor Pressure (mbar) \\ 
Wind Speed (m/s) & Atmospheric Pressure (mbar) \\ 
Wind Direction ($^{\circ}$) & Vapor Pressure Deficit (mbar) \\ 
Precipitation (mm) & Relative Humidity (\%) \\ 
Max Precipitation Rate (mm/h) & Specific Humidity (g/kg) \\ 
Solar Radiation (W/m$^2$) & Concentration of H$_2$O (mmol/mole) \\ 
Lightning Activity & Air Density (g/m$^3$) \\ 
Lightning Distance (km) & Wind Velocity (m/s) \\ 
Logger Temperature ($^{\circ}$C) & Wind Direction ($^{\circ}$) \\ 
Battery Voltage (mV) & Maximum Wind Velocity (m/s) \\ 
Battery Percent (\%) &  Rainfall (mm)\\ 
Gust Speed (m/s) & Rainfall Duration (s) \\ 
RH Sensor Temperature ($^{\circ}$C) & Shortwave Downward Radiation (W/m$^2$) \\ 
Soil Temperature ($^{\circ}$C) & Photosynthetically Active Radiation ($\mu$mol/m$^2$/s) \\ 
Leaf Wetness (min) & CO$_2$ (ppm) \\ 
\hline
\end{tabular}
\end{table}

\end{document}